\documentclass[11pt,a4paper]{article}
\usepackage[hyperref]{acl2020}
\usepackage{times}
\usepackage{latexsym}

\usepackage{microtype}
\usepackage{amssymb}
\usepackage{graphicx}

\usepackage{url}
\usepackage{amsfonts}
\usepackage{amsmath}
\usepackage{graphicx}
\usepackage{hyperref}
\usepackage{subcaption}
\usepackage{cleveref}
\usepackage{balance}

\aclfinalcopy % Uncomment this line for the final submission

\newcommand{\ie}{\textit{i}.\textit{e}.}
\newcommand{\eg}{\textit{e}.\textit{g}.}

\newcommand{\secref}[1]{\S\ref{#1}}

\newcommand{\cat}{%
  \begingroup\normalfont
  \includegraphics[height=1.5\fontcharht\font`\B]{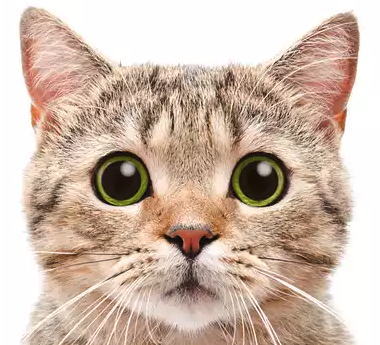}%
  \endgroup
}
\newcommand{\catx}{%
  \begingroup\normalfont
  \includegraphics[height=1.5\fontcharht\font`\B]{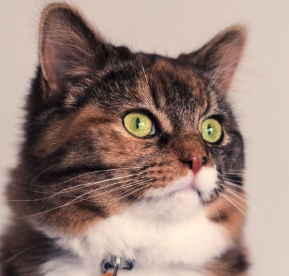}%
  \endgroup
}

\title{Multilingual Multimodal Pre-training for \\ Zero-Shot Cross-Lingual Transfer of Vision-Language Models}

\author{Po-Yao Huang$^{13}$\thanks{~~Equal contribution.}, Mandela Patrick$
^{23}$\footnotemark[1], Junjie Hu$^1$, 
\\
\textbf{Graham Neubig}$^1$, \textbf{Florian Metze}$^3$, \textbf{Alexander Hauptmann}$^1$ \\
  $^1$School of Computer Science, Carnegie Mellon University \\
  $^2$Visual Geometry Group, University of Oxford \\
  $^3$Facebook AI \\ 
      \{poyaoh,junjieh,gneubig,alex\}@cs.cmu.edu, 
      \{mandelapatrick,fmetze\}@fb.com}
\date{}

\begin{document}
\maketitle
\begin{abstract}
This paper studies zero-shot cross-lingual transfer of vision-language models.
%First, we perform an empirical study on cross-lingual transfer of text-video models in the zero-shot setting.
%Despite the success in the NLP domain, we observe that text-to-video search models with multilingual Transformer perform significantly worse with non-English queries.
%\gn{This sentence is hard to parse. What does ``Even with multilingual Transformers'' mean in the context of ``fine-tuned English-vision'' models? Is ``Transformers'' really important, could you be more general? It would be good to stress the contribution, such as starting the sentence with ``First, we perform an empirical study that demonstrates''.}.
%
%
Specifically, we focus on multilingual text-to-video search and propose a Transformer-based model that learns contextualized multilingual multimodal embeddings.
%
%Compared with English queries, our empirical study demonstrates significant performance degradation with non-English queries for text-to-video search under a zero-shot setting.
Under a zero-shot setting,
we empirically demonstrate that performance degrades significantly when we query the multilingual text-video model with non-English sentences.
%
%We empirically demonstrate that performance degrades significantly when such model are used used with non-English queries for text-to-video search under a zero-shot setting.
%We first demonstrate empirically that performance degrades significantly when such models are used with non-English queries for text-to-video search under a zero-shot setting.  
% revised and changed to the version above-Bernie
%
%
To address this problem, we introduce a multilingual multimodal pre-training strategy, and collect a new multilingual instructional video dataset (Multi-HowTo100M) for pre-training.
Experiments on VTT show that our method significantly improves video search in non-English languages without additional annotations. 
Furthermore, when multilingual annotations are available, our method outperforms recent baselines by a large margin in multilingual text-to-video search on VTT and VATEX; as well as in multilingual text-to-image search on Multi30K.
Our model and Multi-HowTo100M is available at \url{http://github.com/berniebear/Multi-HT100M}

\end{abstract}

\section{Introduction}
One of the key challenges at the intersection of computer vision (CV) and natural language processing (NLP) is building versatile vision-language models that not only work in English, but in all of the world’s approximately 7,000 languages.
Since collecting and annotating task-specific parallel multimodal data in all languages is impractical, a framework that makes vision-language models generalize across languages is highly desirable.

%% Paragraph 2: Solution in the NLP domain
One technique that has shown promise to greatly improve the applicability of NLP models to new languages is \emph{zero-shot cross-lingual transfer}, where models trained on a source language are applied as-is to a different language without any additional annotated training data \citep{tackstrom-etal-2012-cross,klementiev2012inducing,cotterell-heigold-2017-cross,chen2018adversarial,neubig-hu-2018-rapid}.
In particular, recent techniques for cross-lingual transfer have demonstrated that by performing unsupervised learning of language or translation models on many languages, followed by downstream task fine-tuning using only English annotation, models can nonetheless generalize to a non-English language
%without collecting additional annotations in that language % -> duplicate with previous sent.
\citep{wu2019beto,lample-conneau,huang2019unicoder,artetxe-etal-2020-cross,xtreme}.
This success is attributed to the fact that many languages share a considerable amount of underlying vocabulary or structure. 
At the vocabulary level, languages often have words that stem from the same origin, for instance, ``desk'' in English and ``Tisch'' in German both come from the Latin ``discus''.
At the structural level, all languages have a recursive structure, and many share traits of morphology or word order.

%\bernie{I think the transition of this paragraph is a bit weird and made an (imperfect) revision. Inputs welcomed 5:30pm}

%% Paragraph 3.0: How to generalize cross-lingual transfer of VL models 
%% Paragraph 3.1: why potentially 3.0 will work 

For cross-lingual transfer of vision-language models, the visual information is clearly an essential element. 
To this end, we make an important yet under-explored step to incorporate visual-textual relationships for improving multilingual models~\cite{devlin-etal-2019-bert,artetxe-etal-2020-cross}.
While spoken languages could be different, all humans share similar vision systems, and many visual concepts 
%in the world% % save space
can be understood universally~\cite{sigurdsson2020visual, Zhang2020Neural}.
For example, while \cat{} is termed ``cat'' for an English speaker and ``chat'' for a French speaker; 
%and ``mao'' for a Chinese speaker,
they understand \cat{} similarly.
We leverage this observation to learn to associate sentences in different languages with visual concepts for promoting cross-lingual transfer of vision-language models.

%% Paragraph 4:
%% 1. What we actually implemented: video-text Transformer
%% 2. what problem we faced: not good for zero-shot
%% 3. How we resolve it: MMP
In this work, we focus on multilingual text-to-video search tasks and propose a Transformer-based video-text model 
%with tailored inter-modal, intra-modal, and cross-lingual contrastive objectives 
to learn contextual multilingual multimodal representations.
Our vanilla model yields state-of-the-art performance in multilingual text$\rightarrow$video search when trained with multilingual annotations.
However, under the zero-shot setting, rather surprisingly, there is a significant performance gap between English and non-English queries (see \secref{sec:exp:cross_lingual} for details).
To resolve this problem, 
motivated by recent advances in large-scale language model~\cite{artetxe-etal-2020-cross} and multimodal pre-training~\cite{vilbert,ht100m,Patrick2020MultimodalSF},
we propose a multilingual multimodal pre-training (MMP) strategy  %\gn{what are the novel characteristics of this training strategy compared to related past work? just a few words is fine.}.
to exploit the weak supervision from large-scale multilingual text-video data.
We construct the Multilingual-HowTo100M dataset, that extends the English HowTo100M~\cite{ht100m} dataset to contain subtitles in 9 languages for 1.2 million instructional videos.

%% Paragraph 5: Why MMP works
%% Intuitions and advantages of the proposed solution:
%% 1. Data augmentation: improve robustness for English-Video (like XLDA (cite XLDA.) => May not be appreciated by language community 
%% 2. Learning the structure shared across languages. Explain the reason why multilingual > monolingual
%% 3. Enable/Improve zero-shot cross-lingual text-video tasks (Table4)

Our method has two important benefits. % Take (2) and (3)
First, compared to pre-training on English-video data only, pre-training on multilingual text-video data 
exploits the additional supervision from a variety of languages, %underlying text similarities across languages,
and therefore, enhances the search performance on an individual language.
%\gn{This is a bit vague. Is it because there is more training data when you use data from all languages? If not additional training \emph{data} (with respect to the number of videos), then maybe it's additional training \emph{signal} by having to learn for more different languages? Again just a few words of clarification is probably OK.}.
%\bernie{Both (multi-task and more data))
Second, 
by exploiting the visual data as an implicit ``pivot'' at scale, our methods learns better alignments in the multilingual multimodal embedding space (\eg, ``cat''-\cat{}-``chat''), which leads to improvement in zero-shot cross-lingual transfer (\eg, from ``cat''-\cat{} to ``chat''-\catx{}) of vision-language models.
%\gn{not exactly sure what ``structured'' means here? maybe just delete, but you could also clarify.}

% View as data augmentation
%First, pre-training with multilingual data can be viewed as a data augmentation strategy, which has been shown effective in the ~\cite{xlda, autoaugment, Cubuk2019RandAugmentPD, yun2019cutmix} and audio domain~\cite{Park2019SpecAugmentAS}. Consequentially, this has also been confirmed with our finding that balabala.

%% Paragraph 6: Experimental highlights 
%% remark: add \% diff of XLM-R in NLP ?
%We conduct extensive experiments to verify these benefits.
In our experiments on VTT~\cite{vtt} and VATEX~\cite{vatex}, our method yields state-of-the-art English$\rightarrow$video search performance. 
For zero-shot cross-lingual transfer, the proposed multilingual multimodal pre-training improves English-video pre-training by $2 \sim 2.5$ in average R@1 across 9 languages.
Additionally, when trained with in-domain multilingual annotations as other baselines, our method outperforms them by a large margin in multilingual text$\rightarrow$video search on VATEX and text$\rightarrow$image search on Multi30K~\cite{Multi30K}.

%% Paragraph 7: Summary of contributions
To summarize, we make the following contributions: 
(1) We propose a transformer-based video-text model that learns contextual multilingual multimodal representations (\secref{sec:model}). 
% (2) We experiment on testing the zero-shot cross-lingual transferabilities of vision-language models, which demonstrates that the results are notably worse than expected compared to transferring text-only model in the NLP domain. 
(2) We empirically demonstrate that vision-language models, unlike NLP models, have limited zero-shot cross-lingual transferrability.
(\secref{sec:exp:cross_lingual}).
% (3) To resolve this problem, we introduce the multilingual multimodal pre-training strategy and construct a new Multi-HowTo100M dataset for pre-training (\secref{sec:mutliHT100M}).
(3) We introduce the multilingual multimodal pre-training strategy and construct a new Multi-HowTo100M dataset (\secref{sec:mutliHT100M}) for pre-training to improve zero-shot cross-lingual capability of vision-language models.
(4) We demonstrate the effectiveness of our approach, by achieving state-of-the-art multilingual text$\rightarrow$video search performance in both the zero-shot (\secref{sec:exp:cross_lingual}) and fully supervised setup (\secref{sec:exp:sota}).
%(\secref{sec:exp:sota} -\secref{sec:exp:img_text}).

%\gn{Experiments on testing the cross-lingual generalization abilities of vision-text models, which demonstrate that results are notably worse than you may expect by looking at the performance of transfer of models in the text-only domain

\section{Related Work}
\textbf{Cross-lingual representations.}
Early work on
learning non-contextual cross-lingual representations used either parallel
corpora~\cite{gouws-sogaard-2015-simple, luong-etal-2015-effective} or a bilingual dictionary to learn a transformation~\cite{faruqui-dyer-2014-improving, mikolov2013efficient}. 
Later approaches reduced the amount of supervision using self-training~\cite{artetxe-etal-2017-learning}.
%and unsupervised strategies such as adversarial training~\cite{conneau2017word} and heuristic initialisation~\cite{artetxe2018robust}.
With the advances in monolingual transfer learning~\cite{mccaan, howard-ruder-2018-universal, peters-etal-2018-deep, devlin-etal-2019-bert}, multilingual extensions of pre-trained encoders have been proven effective in learning deep contextual cross-lingual representations~\cite{eriguchi-etal-2017-learning, lample-conneau, wu-dredze-2019-beto, siddhant2020leveraging, pires-etal-2019-multilingual, madx}. 
We extend prior work to incorporate visual context.

%Large-scale training data has enabled effective pre-training of image~\citep{yalniz2019billionscale, Sun_2017}, video~\citep{ghadiyaram2019large, thomee2016yfcc100m} and textual representations~\citep{raffel2019exploring}.
\noindent\textbf{Video-text representations.}
The HowTo100M dataset~\cite{ht100m} has attracted significant interest in leveraging multimodal pre-training for text$\rightarrow$video search~\citep{korbar2020video},  captioning~\citep{iashin2020multi}, and unsupervised translation via image-based~\citep{Globetrotter, ummt} and video-based~\citep{sigurdsson2020visual} alignment.
This work studies a challenging and unexplored task: Zero-shot cross-lingual transfer of vision-language models.
Unlike prior image/video-text work that utilizes RNN~\cite{dual,hgr,smalr,mule} and inter-modal contrastive objectives~\cite{sigurdsson2020visual,liu2019use,mhad,support}, we employ Transformers to learn contextual multilingual multimodal representations and uniquely models cross-lingual instances.
Moreover, we build Multi-HowTo100M, the largest text-video dataset for multilingual multimodal pre-training.

%Although semantically rich and diverse, instructional videos from the web are super \gn{``very'' (``super'' is too informal for a paper)} noisy and therefore several approaches have been proposed to combat this.
%A few works~\citep{Sun_2019, zhu2020actbert, sun2019learning} extend the BERT model to accept both visual and textual tokens to learn high-level semantic video-text representations.

\begin{figure*}[t!]
    \centering
    \includegraphics[width=1.0\linewidth]{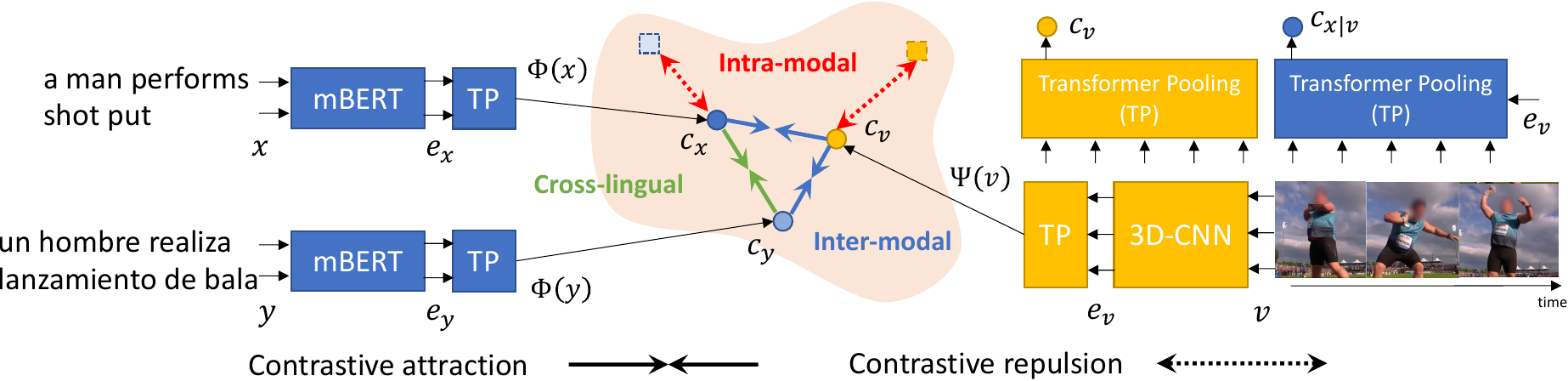}
    \caption{An overview of our video-text model for learning contextual multilingual multimodal representations. We utilize \textit{intra-modal}, \textit{inter-modal}, and conditional \textit{ cross-lingual} contrastive objectives to align $(x, v, y$) where $x$ and $y$ are the captions or transcriptions in different languages of a video $v$. TP: Transformer pooling head.
    }\label{fig:model}
\end{figure*}

\noindent\textbf{Cross-lingual Transfer.}
Cross-lingual transfer has proven effective in many NLP tasks including dependency parsing~\cite{schuster-etal-2019-cross}, 
named entity recognition~\cite{rahimi-etal-2019-massively}, 
sentiment analysis~\cite{barnes-etal-2019-sentiment}, 
%natural language inference~\cite{conneau-etal-2018-xnli}, 
document classification~\cite{SCHWENK18_658}, 
and question answering~\cite{lewis-etal-2020-mlqa, artetxe-etal-2020-cross}. 
%Evaluation on a single task is problematic as past work has noted potential issues with standard datasets: 
%MLDoc~\cite{SCHWENK18_658} can be solved by matching keywords~\cite{artetxe-etal-2020-call}, while MultiNLI, the dataset from which XNLI~\cite{conneau-etal-2018-xnli} was derived, contains superficial cues that can be exploited~\cite{gururangan-etal-2018-annotation}. 
%Evaluation on multiple tasks is thus necessary to fairly compare cross-lingual models. 
%Benchmarks covering multiple tasks like GLUE~\cite{wang-etal-2018-glue} and SuperGLUE~\cite{wang2020superglue} have arguably spurred research in monolingual transfer learning. 
%In the cross-lingual setting, such a benchmark not only needs to cover a diverse set of tasks, but also languages. 
Recently, XTREME~\cite{xtreme} was proposed to evaluate the cross-lingual transfer capabilities of multilingual representations across a diverse set of NLP tasks and languages. However, a comprehensive evaluation of multilingual multimodal models on zero-shot cross-lingual transfer capabilities is still missing.
%In this work, we make make the attempt to study the feasibility of transferring V-L models across languages in the zero-shot setting
To our best knowledge, we are the first work that investigates and improves zero-shot cross-lingual transfer of vision-language models.
%and makes attempt improving the baseline with multilingual Transformers.

%Figure~\ref{fig:model} gives an illustration of our model architecture.

\section{Method\label{sec:method}}
We consider the problem of learning multilingual multimodal representations from a corpus $\mathcal{C}$ of video-text pairs $\{(x_i,v_i)\}_{i=1}^{C}$, where $v_i$ is a video clip and $x_i$ is its corresponding text (caption or transcription) that is written in one of $K$ languages.
Our goal is to learn a shared multilingual text encoder $c_x = \Phi(x)$ and a video encoder $c_v = \Psi(v)$, both of which project the input to a shared $D$-dimensional embedding space $c_v,c_t\in\mathbb{R}^D$, where semantically similar instances (\ie, paired $(x_i, v_i)$) are closer to each other than the dissimilar ones (\ie, $(x_i, v_j), i \neq j$).
In the following, we denote a batch of multilingual text-video samples as $\mathcal{B}=\{(x_i,v_i)\}_{i=1}^{B}\}$ 
where $\mathcal{B}\subset\mathcal{C}$. 

%\gn{The second ``B'' here should not be in ``mathcal'', right?} => yes

\subsection{Multilingual Multimodal Transformers\label{sec:model}} 
Figure~\ref{fig:model} gives an overview of the proposed method.
Our text encoder consists of a multilingual Transformer (\eg~multilingual BERT~\cite{devlin-etal-2019-bert}) and a text Transformer pooling head (explained below).
Similarly, our video encoder consists of a 3D-CNN (\eg~R(2+1)D network~\cite{tran2018closer}) and a video Transformer pooling head. 
We use these multilingual multimodal Transformers to encode text and video for alignment.

%Built upon multilingual Transformer (\eg~multilingual BERT (mBERT; \citet{bert})) and 3D-CNN (\eg~R(2+1D) network~\cite{tran2018closer}), we propose to use multilingual multimodal Transformers to encode text and video for alignment. 
%\gn{Here ``transformers'' is probably important, because you're discussing the specific architecture.} 

%,built upon pretrained multilingual Transformers (\ie~multilingual BERT (mBERT; \citet{bert}) or XLM-Roberta (XLM-R; \citet{artetxe-etal-2020-cross})),

Unlike prior multilingual text-image models~\cite{gella2017image,mule,mhad} that utilize word embeddings and RNNs,
our multilingual text encoder is built on a multilingual Transformer that
generates contextual multilingual representations $e_x\in\mathbb{R}^{N\times D}$ to encode a sentence $x$ containing $N$ words.
We employ an additional 2-layer Transformer which we will call a ``Transformer pooling head (TP)'' as it serves as a pooling function to selectively encode variable-length sentences and aligns them with the corresponding visual content. 
We use the first output token of the second Transformer layer as the final sentence representation.
%Precisely, we set
%$
%c_x = \Phi(x) =
%(T_x^2 \circ T_x^1(e_x))[0]
%$,
%where
%$
%T_x^i(e_x)
%=
%\text{LN}(\text{FFN}(M(e_x))+M(e_x)) 
%$ and $M(e_x) = \text{LN}(e_x+\text{MHA}(e_x))$
%is the $i$-th transformer layer with multi-head self-attention (MHA), %layer-normalization (LN), and feed-forward network (FFN).
Precisely, we set $c_x=\text{Trans}_x^{(2)}(\text{query=key=value=}e_x)[0]$ where $\text{Trans}_x^{(2)}$ is a 2-layer stack of  Transformers~\citep{vaswani2017attention} with $e_x$ as the (\text{query},\text{key},\text{value}) in the multi-head attention. 
Note that we use the same text encoder to encode sentences in all languages.

For encoding videos, our model uses pre-trained 3D-CNNs that encode spatial-temporal context in a video.
For a $M$-second video $v$, we apply R(2+1)D~\cite{tran2018closer} and S3D~\cite{miech19endtoend} networks to its frames, concatenate network outputs, and apply a linear layer to encode the visual input, $e_v\in\mathbb{R}^{M\times D}$, to our model.
% A video is encoded as .
% too many details
%where we additionally concatenate a $D$-dimension vector that contains the maximum of features across $M$ seconds in the first column that serves as a clip-level overview.
%\gn{Very long run-on sentence here, would be good to break it up into one or two.}. 
Similarly to the text part, we employ a two-layer Transformer as the pooling head to encode videos with different lengths into fixed-length representations.
%Formally, we set
%$
%c_v = \Psi(v) =
%(T_v^1 \circ T_v^2(e_v))[0]
%$.
Formally, we set 
$c_v=\text{Trans}_v^{(2)}(\text{query=key=value=}e_v)[0]$.
Since videos are typically long and have a high frame rate (\eg, 30 fps), it is infeasible to update 3D-CNNs simultaneously and therefore, we use pre-extracted video features. Our model is parameterized by $\theta=\theta_{\text{mBERT}} \cup \theta_{\text{Trans}_x} \cup \theta_{\text{Trans}_v}$.

\subsection{Multilingual Text-Video Alignment}
\label{sec:multi_align}

%\gn{Somewhat important: I don't think the objectives you write below are actually what is normally considered ``noise contrastive estimation'', which is a specific unbiased estimator for probabilistic models \citep{gutmann2010noise}. What you describe below is probably just a ``contrastive loss'', right?}
%\bernie{It is and I will try clarify it with ref. The noise distribution is, either sampled from noised view of text-text or video-video, or from,  inter-modal (considerd as) a noised-view.}

For learning multimodal representations, the common practice is to minimize a contrastive objective to map the associated (video, text) embeddings to be near to each other in a shared embedding space.
The inter-modal max-margin triplet loss has been widely studied in video-text~\cite{Yu_2018,liu2019use} and image-text~\cite{mule,smalr,mhad} research.
In this work, we generalize and model all \textit{inter-modal}, \textit{intra-modal}, and \textit{cross-lingual} instances with a noise contrastive estimation objective
(NCE)~\cite{gutmann2010noise, oord2018representation, Chen2020ASF}.

%\begin{equation}\label{e:contrast-objective}
%\mathcal{L}(\mathcal{X},\mathcal{V})
%=-\frac{1}{B}\sum_{i=1}^B
%\text{log} \frac{e^{\frac{1}{\tau}s(\Phi(x_i), \Psi(v_i))}}
%{\sum_{j=1}^{B} e^{\frac{1}{\tau}s(\Phi(x_j), \Psi(v_j))}}
%\end{equation}

\noindent \textbf{Inter-modal NCE.}
%\gn{``contrastive'' Make sure that the paper is typo-free before you submit.}
Let $\mathcal{X}$ and $\mathcal{V}$ denote the subsets of the sampled sentences in multiple languages and videos in $\mathcal{B}$, respectively. 
And let $s(a,b)=\frac{a^Tb}{\|a\|\|b\|}$ be the cosine similarity measure. We use an (inter-modal) NCE objective defined as:
\begin{equation}\label{e:inter}
\mathcal{L}(\mathcal{X},\mathcal{V})=
-\frac{1}{B}\sum_{i=1}^B\text{log}
\ell^{\text{NCE}}(\Phi(x_i),\Psi(v_i)),
\end{equation}
where
\begin{equation}\label{e:contrast-objective}
\ell^{\text{NCE}}(c_x,c_v)=
\frac{e^{s(c_x, c_v)}}
{e^{s(c_x, c_v)} +
%\sum\limits_{(x',v')\sim\mathcal{N}}e^{s(x, v)}},
\sum_{(x',v')\sim\mathcal{N}}e^{s(c_{x'}, c_{v'})}}
\end{equation}

\noindent In inter-modal NCE, $\mathcal{L}^{\text{inter}}=
\mathcal{L}(\mathcal{X},\mathcal{V})$, the noise $\mathcal{N}$ is a set of ``negative'' video-text pairs sampled to enforce the similarity of paired ones are high and and those do not are low.
Following~\citet{miech19endtoend}, we set the negatives of $(x_i,v_i)$ as other $x_j$ and $v_j, j \neq i$ in $\mathcal{B}$.

Intuitively, inter-modal NCE draws paired (semantically similar) instances closer and pushes apart non-paired (dissimilar) instances. 
Note that we do not distinguish language types in $\mathcal{X}$ and the sentences in all possible languages will be drawn towards their corresponding videos in the shared multilingual text-video embedding space.

\noindent \textbf{Intra-modal NCE.}
Beyond cross-modality matching, we leverage the intra-modal contrastive objective to learn and preserve the underlying structure within the video and text modality. 
For example, \textit{Corgi} should be closer to \textit{Husky} than \textit{Balinese}. 
Prior image-text work~\cite{gella2017image,ann} utilizes a triplet loss to maintain such neighborhood relationships.
%\gn{Normally augmentation makes something bigger, more correct, or more extensive, where in fact here you're doing the opposite. I changed ``augmented version'' to ``noised version'', please check that you think that's OK.} 
% bernie: sounds good
Inspired by recent success in self-supervised image and video representation learning~\cite{yalniz2019billionscale,ghadiyaram2019large}, our model leverages intra-modal NCE 
%to draw a video/text towards a noised version of itself and to push away from other instances in the same modality.
that constrains the learned representations to be invariant against noise and to maintain the within-modality structure simultaneously.
We minimize the following intra-modal NCE loss:
\begin{equation}
\mathcal{L}^{\text{intra}}=
\mathcal{L}(\mathcal{X},\mathcal{X}^m)
+
\mathcal{L}(\mathcal{V},\mathcal{V}^m),
\label{e:intra}
\end{equation}
where $\mathcal{X}^m$ and $\mathcal{V}^m$ are the noised version of the original sentences and videos. 
For noising, we randomly mask 5\% of the multilingual text tokens and video clips.
We optimize our model by
\begin{equation}
\min_{\theta} 
\mathcal{L}^{\text{inter}}
+
\mathcal{L}^{\text{intra}}
\label{e:full-objective}
\end{equation}

\subsection{When Visually-Pivoted Multilingual Annotations Are Available}
\label{sec:pivot_align}

%\gn{This subsection is the only one where the title is not in title case. Probably better to make it title case.}

%(five (\textit{en},\textit{zh}) translation pairs; and five non-paired \textit{en} and \textit{zh} descriptions).

In many multilingual multimodal datasets,
there are sentences in different languages that describe a shared visual context. 
For example, 10 English and 10 Chinese descriptions are available for each video in VATEX.
With these visually-pivoted (weakly paralleled) sentences $(x,y)$, we further revise the contrastive objectives to leverage this additional supervisory signal. 
Given a visually-pivoted corpus $\mathcal{C}^p$ that contains all possible combination of visually-pivoted pairs  $\{(x_i,v_i,y_i)\}_{i=0}^{C_p}$, we sample batches $\mathcal{B}^p=\{(x_i,v_i,y_i)\}_{i=1}^{\mathcal{B}^p},  \mathcal{B}^p\subset\mathcal{C}^p$ and revise the contrastive objective as:
\begin{align}
\mathcal{L}^{\text{inter}}&=
\mathcal{L}(\mathcal{X},\mathcal{V})
+
\mathcal{L}(\mathcal{Y},\mathcal{V})
\label{e:inter2}
\\
\mathcal{L}^{\text{intra}}&=
\mathcal{L}(\mathcal{X},\mathcal{X}^m)
+
\mathcal{L}(\mathcal{Y},\mathcal{Y}^m)
+
\mathcal{L}(\mathcal{V},\mathcal{V}^m)
\label{e:intra2}
\end{align}

\noindent \textbf{Visual-pivoted Cross-lingual NCE.}
Inspired by Translation Language Modeling (TLM) in XLM~\cite{lample-conneau},
we propose a multimodal TLM-like contrastive objective which promotes alignments of descriptions in different languages that describe the same video.
We use the intuition that conditioned on a video, the  descriptions (need not to be translation pairs) in different languages would likely be semantically similar. To this end, we set the cross-lingual NCE as:

\begin{equation}
\mathcal{L}^{\text{cross}}=
\mathcal{L}(\mathcal{X}|\mathcal{V},\mathcal{Y}|\mathcal{V})\label{e:cross}
\end{equation}

For visually-pivoted sentences, as shown in Fig.~\ref{fig:model}, we generate their representations conditioned on the video they describe. 
We extend the \textit{key} and \textit{value} of multihead attention with the additional visual content $e_v$ and generate new $c_{x|v}$ and $c_{y|v}$ for matching. 
Specifically, our model employs
$c_{x|v}=\text{Trans}_x^{(2)}(\text{query=}e_x, \text{key=value=}e_x||e_v)[0]$.
With the access to (visually-pivoted) multilingual annotations, we optimize our model by

\begin{equation}
\min_{\theta} 
\mathcal{L}^{\text{inter}}
+
\mathcal{L}^{\text{intra}}
+
\mathcal{L}^{\text{cross}}
\end{equation}

\begin{figure*}
    \centering
    \includegraphics[width=1.0\linewidth]{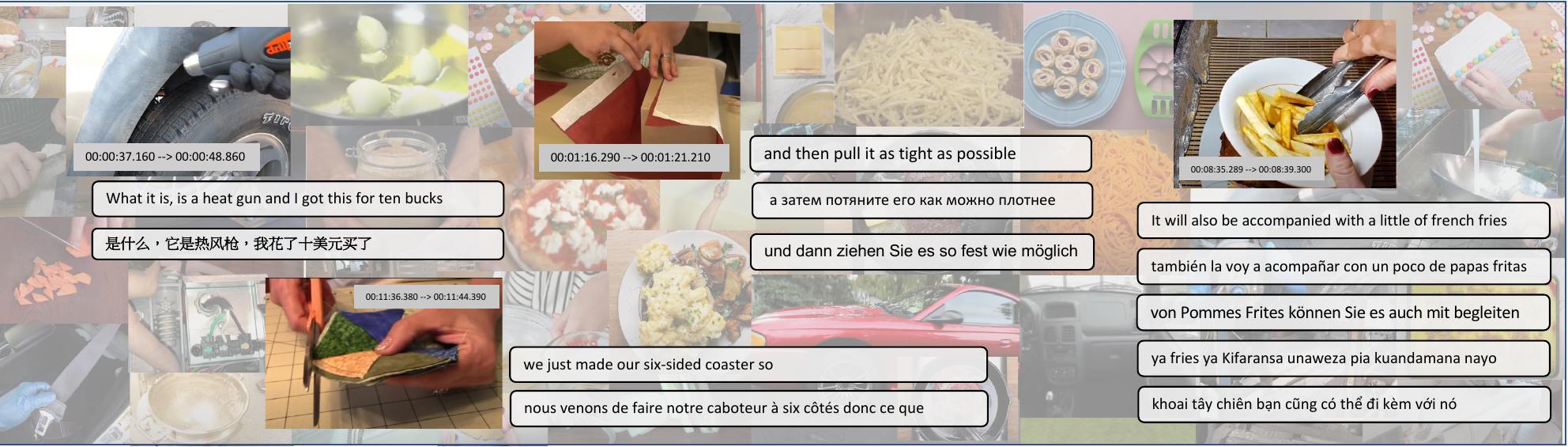}
    \caption{Video clips and the corresponding multilingual subtitles in Multi-HowTo100M.}\label{fig:dataset_vis}
\end{figure*}

At the inference time, we simply apply $c_x=\Phi(x)$ and $c_v=\Psi(v)$ to encode multilingual text queries and videos.
For text-to-video search, we sort videos according to their cosine similarity scores to the text query.

\section{The Multilingual HowTo100M Dataset\label{sec:mutliHT100M}}
As large-scale pre-training has been shown important in recent NLP and vision-language models,
we construct the~\textbf{Multilingual HowTo100M} dataset (Multi-HowTo100M) to facilitate research in multilingual multimodal learning.
The original HowTo100M~\citep{ht100m} dataset is a large-scale video collection of 1.2 million instructional videos (around 138 million clips/segments) on YouTube, along with their automatic speech recognition (ASR) transcriptions as the subtitles. 
For each video in HowTo100M, 
we crawl and collect the multilingual subtitles provided by YouTube, which either consist of user-generated subtitles or those generated by Google ASR and Translate in the absence of user-generated ones.
Essentially, we collect video subtitles in 9 languages: English (\textit{en}), German (\textit{de}), French (\textit{fr}), Russian (\textit{ru}), Spanish (\textit{es}), Czech (\textit{cz}), Swahili (\textit{sw}), Chinese (\textit{zh}), Vietnamese (\textit{vi}).

%\gn{It's not clear here what you mean by ``user-generated subtitles and translated subtitles'', do you mean ``user generated translations and Google translations''?}
% Icelandic (\textit{is}).
%\bernie{yes, fan will translate the caption into other languages. The most of multilingual caption are generate by YouTube(Google) ASR + Google translation }
%\gn{Maybe make this more clear, and also make it clear whether you're using both when the fan translated ones exist, or if you're using Google as a fallback when they don't exist. For example, you could also write ``We use the subtitles provided by YouTube, which either consist of user-generated subtitles or those generated by Google ASR and Translate in the absence of user-generated ones.'' (not sure if this is correct, but that's the level of specificity that I'm thinking about)}
%\bernie{OK, sounds clear. I will revise it}

At the time of dataset collection (May 2020), there are 1.1 million videos available, each with subtitles in 7-9 languages.
%(\ie~69-70 million  ASR transcriptions). 
The video length ranges from 1 minute to more than 20 minutes.
We utilize Multi-HowTo100M for multilingual multimodal pre-training to exploit the weak supervision from large-scale multilingual text-video data.
In Fig.~\ref{fig:dataset_vis}, we provide a visualization of few instances sampled in Multi-HowTo100M with the corresponding video frame, timestamp, and transcriptions in different languages.
Please refer to Appendix for more details and dataset statistics.

\section{Experiment\label{sec:exp}}

In this section, we first describe our experimental setup (\S\ref{sec:exp:dataset}-\ref{sec:exp:setup}). 
In \S\ref{sec:exp:albation}, we conduct ablation studies to validate the effectiveness of proposed multilingual text-video model . 
With the best models at hand, we investigate their zero-shot cross-lingual transferability in \S\ref{sec:exp:cross_lingual}, where we showcase that the proposed multilingual multimodal pre-training serves as the key facilitator. 
We then verify the superior text$\rightarrow$video search performance of our method under the
monolingual, multilingual, and cross-modality settings in \S\ref{sec:exp:img_text}.

\subsection{Evaluation Datasets\label{sec:exp:dataset}}
%VTT
\noindent\textbf{MSR-VTT} (VTT)~\citep{vtt} contains 10K videos, where each video is annotated with 20 captions.
Additionally, we created pseudo-multilingual data by translating the English captions into 8 languages with off-the-shelf machine translation models.\footnote{https://marian-nmt.github.io/} %\gn{I put ``pseudo-'' here because I think it's important that we at least signal that this is not actual multilingual data. Creating data through translation can cause biased experimental results (as Junjie knows well from XTREME).} => OK
We use the official training set (6.5K videos) and validation set (497 videos). 
We follow the protocol in~\citet{ht100m, liu2019use} which evaluates on text$\rightarrow$video search with the 1K testing set defined by~\citet{Yu_2018}. 

%VATEX
\noindent\textbf{VATEX}~\citep{vatex} is a multilingual (Chinese and English) video-text dataset with 35K videos. 
%There are 10 English and 10 Chinese descriptions available for each video.
Five (\textit{en},\textit{zh}) translation pairs and five non-paired \textit{en} and \textit{zh} descriptions are available for each video.
We use the official training split (26K videos) and follow the testing protocol in~\citet{hgr} to split the validation set equally into 1.5K validation and 1.5K testing videos. 

\noindent\textbf{Multi30K}~\cite{Multi30K} is a multilingual extension of Flickr30K~\cite{Flickr30K}. 
For each image, there are two types of annotations available: (1) One parallel (English,German,French,Czech) translation pair and (2) five English and five German descriptions collected independently. 
The training, validation, and testing splits contain 29K, 1K, and 1K images respectively.

% move to exp setup 
%We only use the English annotations for the zero-shot experiments and use all available annotations for benchmarking multilingual multimodal representations with other recent work.

\subsection{Implementation Details\label{sec:exp:implementation}}

For the video backbone, we use a 34-layer, R(2+1)-D~\citep{tran2018closer} network pre-trained on IG65M~\cite{ghadiyaram2019large} and a S3D~\cite{miech19endtoend} network pre-trained on HowTo100M.
We pre-extract video features and concatenate the two 3D-CNN outputs to form $e_x \in \mathbb{R}^{M\times1024}$ as a video input.

%\gn{This isn't true any more, there are much better models on the XTREME leaderboard XLM-R is 68.2 and T-ULRv2 + StableTune is 80.7 average accuracy.}.
For the text backbone, we use multilingual BERT (mBERT)~\cite{devlin-etal-2019-bert} or XLM-Roberta-large (XLM-R)~\cite{artetxe-etal-2020-cross}, where the latter achieves near SoTA zero-shot cross-lingual transfer performance for NLP tasks. 
Following~\citet{xtreme}, instead of using the top layer, we output the 12-th layer in XLM-R and mBERT. 
For vision-language tasks, we freeze layers below 9 as this setup empirically performs the best.

%The softmax temperature in NCE is set to $0.1$. => appendix
Our model employs a 2-layer Transformer with 4-head attention for the text and video transformer pooling (TP) modules. The embedding dimension $D$ is set to 1024.
We use the Adam~\citep{kingma2014adam} optimizer and a $0.0002$ learning rate to train our model for 16 (pre-training) and 10 (fine-tuning) epochs.
The softmax temperature in all noise contrastive objectives is set to $0.1$.

\subsection{Experimental Setup\label{sec:exp:setup}}
%{Multilingual Multimodal Pre-training} %\gn{Seems strange to only have one bolded heading in this sub-section} (MMP)
%For models with multilingual multimodal pre-training (MMP), we pre-train on Multi-HowTo100M to learn multilingual multimodal representations.
We use Multi-HowTo100M for multilingual multimodal pre-training (MMP).
For each video, we randomly sample the start and end time to construct a video clip.
For a video clip, we randomly sample one language type each time from 9 languages and use the consecutive ASR transcriptions that are closest in time to compose (text-video) pairs for training.
For simplicity and speed purposes, we follow the training protocol of XLM-R to pre-train on a multilingual corpus \textit{wihtout} using translation pairs,~\ie, we use multilingual text-video pairs $(x, v)$ but no translation pairs from Multi-HowTo100M and utilize only inter- and intra-modal NCE (Eq.~\ref{e:inter}-\ref{e:intra}) for MMP.

We fine-tune our model on VTT, VATEX, and Multi30K to evaluate on text$\rightarrow$video search tasks. 
In the zero-shot cross-lingual transfer experiments, we use only English-video data and fine-tune with Eq.~\ref{e:inter}-\ref{e:intra}. 
We then test the model with non-English queries.
When annotations in additional languages are available
(by humans in VATEX and Multi30K; by MT models (\ie,~\textit{translate-train}) in VTT),
we utilize all available multilingual annotations (\ie, fully supervised) and iterate over all possible $(x,v,y)$ pairs to
train with Eq.~\ref{e:inter2}-\ref{e:cross} to demonstrate the strong performance target for evaluating zero-shot cross-lingual transfer on VTT and to compare fairly with other fully-supervised baselines in multilingual text$\rightarrow$video search on VATEX and Multi30K.
We report the standard recall at $k$ (R@$k$) metrics (higher is better).

\begin{table}[t!]
\small
\centering
      \setlength{\tabcolsep}{2.0pt}
      \begin{tabular}{ccccc}\hline 
      Text-B & Video-B & R@1$\uparrow$ & R@5$\uparrow$ & R@10$\uparrow$\\
      \hline \hline
      XLM-R & S3D & 19.5 & 49.0 & 62.8 \\
      XLM-R & R(2+1)D & 19.0 & 49.5 & 63.2 \\
      XLM-R & R+S & \textbf{21.0} & \textbf{50.6} & \textbf{63.6} \\
      mBERT & R+S & 19.9 & 49.8 & 62.5 \\
      \hline
    \end{tabular}
    \caption{\textbf{Text and Video (B)ackbone comparison.}}\label{tab:feat}
\end{table}

\begin{table}[t!]
\small
\centering
    \setlength{\tabcolsep}{2.0pt}
      \begin{tabular}{ccccc}\hline
      T layers & V layers & R@1$\uparrow$ & R@5$\uparrow$ & R@10$\uparrow$\\
      \hline \hline
      1 & 1 & 20.0 & 50.3 & 63.2 \\
      2 & 1 & 20.1 & 50.5 & 63.8 \\
      2 & 2 & \textbf{21.0} & \textbf{50.6} & 63.6 \\
      $2^*$ & $2^*$ & 20.7 & 50.5 & 63.3 \\ 
      4 & 4 & 20.8 & 50.4 & \textbf{63.8} \\ 
      \hline
    \end{tabular}
    \caption{\textbf{Architecture comparison.} Number of multilingual multimodal transformer layers. \textsuperscript{*}:Weight sharing between video and text transformers.}\label{tab:arch}

\end{table}

\begin{table}[t!]
\small
\centering
    \setlength{\tabcolsep}{2.0pt}
      \begin{tabular}{ccccccc}\hline
      Objective & Inter & Intra & Cross &R@1$\uparrow$ & R@5$\uparrow$ & R@10$\uparrow$\\
      \hline \hline
      Triplet & $\checkmark$& & & 13.3 & 36.0 & 55.2 \\
      Triplet & $\checkmark$& $\checkmark$& &20.9 & 49.3 & 63.0 \\
      NCE & $\checkmark$& & & 21.4  &49.3 & 61.1\\
      NCE & $\checkmark$& $\checkmark$& & 21.0 & 50.6 & 63.6 \\ \hline
      NCE\textsuperscript{*} & $\checkmark$& $\checkmark$ &
       & 21.3 & 50.7 & 63.5 \\
      NCE\textsuperscript{*} & $\checkmark$& $\checkmark$ &
      $\checkmark$ & \textbf{21.5} & \textbf{51.0} & \textbf{63.8} \\ \hline 
      %\hline
    \end{tabular}
    \caption{\textbf{Objective comparison.} *Training with additional machine translated \textit{de}-video and \textit{fr}-video pairs.}\label{tab:obj}

\end{table}

\newcommand{\vv}{{\bf{v}}}%
\newcommand{\ii}{{\bf{i}}}%
\newcommand{\cc}{{$\cdot$}}%

\begin{table*}[th!]
\small
\centering
 \begin{tabular}{lcccccccccc }
      \hline
        Model & $en$ & $de$ & $fr$ & $cs$ & $zh$ & $ru$  & $vi$ & $sw$ & $es$ & Avg$\uparrow$ \\
        \hline \hline 
        mBERT    & 19.9 & 11.1 & 11.6 & 8.2 & 6.9 & 7.9 &  2.7 & 1.4 & 12.0 & 9.1 \\
        mBERT-MP & 20.6 & 11.3 & 11.9 & 8.0 & 7.1 & 7.7 &  2.5 & 1.1 & 12.5 & 9.2 \\
        mBERT-MMP & 21.8 & 15.0 & 15.8 & 11.2 & 8.4 & 11.0  & 3.7  & 3.4  & 15.1 & 11.7 \\
        \hline 
        XLM-R & 21.0 & 16.3 & 17.4 & 16.0 & 14.9 & 15.4 & 7.7 & 5.7 & 17.3 & 14.7\\
        XLM-R-MP & 23.3 & 17.4 & 18.5 & 17.1 & 16.3 & 17.0 & 8.1 & 6.2 & 18.5 & 15.8 \\
        XLM-R-MMP & \textbf{23.8} & \textbf{19.4} & \textbf{20.7} & \textbf{19.3} & \textbf{18.2} & \textbf{19.1} &  \textbf{8.2} & \textbf{8.4} & \textbf{20.4} & \textbf{17.5} \\
        \hline \hline
        mBERT + translated VTT & 19.6 & 18.2 & 18.0 &  16.9 & 16.2 & 16.5 & 8.4 & 13.0 & 18.5 & 16.1
        \\
        mBERT-MMP + translated VTT & 21.5 &  19.1 & 19.8 &  18.3 & 17.3 & 18.3 &  8.9 & 14.1 & 20.0 & 17.4 \\
        XLM-R + translated VTT & 21.5 & 19.6 & 20.1 & 19.3 & 18.9 & 19.1 & 10.3 & 12.5 & 18.9 & 17.8 \\
        XLM-R-MMP + translated VTT & \textbf{23.1} & \textbf{21.1} & \textbf{21.8} & \textbf{20.7} & \textbf{20.0} & \textbf{20.5} & \textbf{10.9} & \textbf{14.4} & \textbf{21.9} & \textbf{19.4}\\
        \hline
    \end{tabular}
    \caption{\textbf{Recall@1 of multilingual text$\rightarrow$video search on VTT.}
    Upper: Zero-shot cross-lingual transfer. Lower: Performance with synthesized pseudo-multilingual annotations for training. 
    MMP: multilingual multimodal pre-training on Multi-HowTo100M.
    MP: Multimodal (English-Video) pre-training on HowTo100M.}\label{tab:cross_lingual}
\end{table*}

\subsection{Comparison Experiments and Ablations\label{sec:exp:albation}}
In this section, we ablate and compare different text/video encoders, Transformer model architectures, and learning objectives for English$\rightarrow$video search on VTT.
%Additional discussion regarding choice of hyperparameters and the Transformer architectures can be found in the supplementary material.

\noindent \textbf{Text and Video Encoders.} 
Table~\ref{tab:feat} compares different text and video encoder backbones. For the visual encoders, while R(2+1)D outperforms S3D, the simple concatenation (\ie, early-fusion) of their output features provides a $1.5 \sim 2.0$ improvement in R@1.
For the text encoder, XLM-R significantly outperforms mBERT.

\noindent \textbf{Transformer Pooling.} 
Table~\ref{tab:arch} compares various configurations of the proposed Transformer pooling module. 
We observe that a simple 2-layer Transformer achieves the best performance. 
Weight sharing of the video and text Transformer slightly degrades the performance. Therefore, we choose to separate them.

\noindent \textbf{Learning Objective.} 
From Table~\ref{tab:obj}, the intra-modal contrastive objective is important for both NCE and Triplet loss. 
In general, the NCE loss outperforms the Triplet loss.
The proposed inter-modal and intra-modal NCE objective achieves the best performance.
When captions in multiple languages are available, cross-lingual NCE additionally provides a consistent improvement.

\begin{figure}[t!]
    \centering
    \subcaptionbox{English$\rightarrow$Video}[0.498\linewidth]{\includegraphics[width=1.0\linewidth]{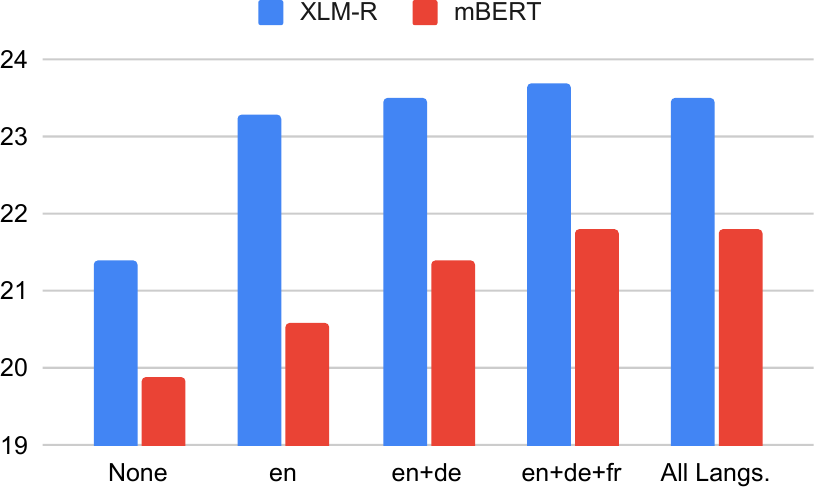}}%
    \subcaptionbox{Zero-shot German$\rightarrow$Video}[0.498\linewidth]{\includegraphics[width=1.0\linewidth]{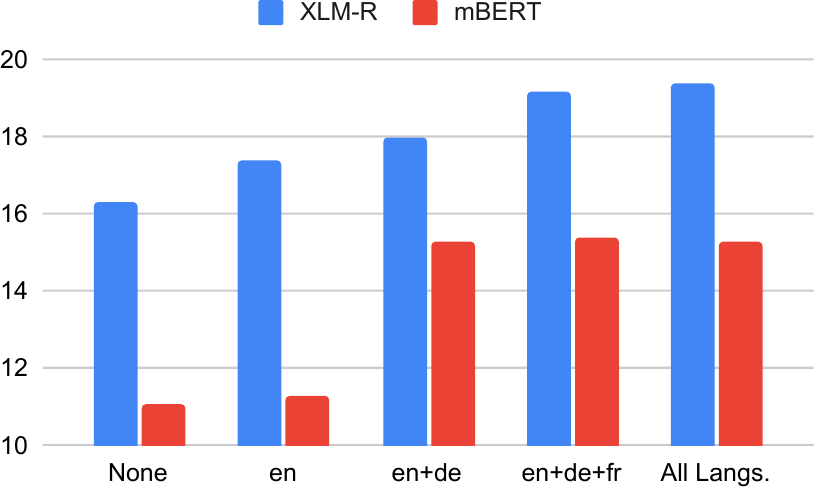}}
    \caption{R@1 trends in languages used for multilingual multimodal pre-training. Left: English$\rightarrow$video search. Right: Zero-shot German$\rightarrow$video search.}\label{fig:trend}
\end{figure}

%%%%%%%%%%%%%%%%%%%%%%%%%%%%%%%%%%
%% Important story of the paper %%
%% clean and careful writing !! %%  
%%%%%%%%%%%%%%%%%%%%%%%%%%%%%%%%%%

% remark: note the terminology pre"-"training, fine"-"tuning, transferrability? 

%\subsection{Effect of Pre-training on Zero-Shot Cross-Lingual Transfer\label{sec:exp:cross_lingual}}
\subsection{VTT Zero-Shot Cross-Lingual Transfer\label{sec:exp:cross_lingual}}

Table~\ref{tab:cross_lingual} shows the multilingual text$\rightarrow$video search results on VTT. 
With the best English-video models at hand (with either mBERT or XLM-R as the text backbone), we first investigate how well these models transfer to other non-English languages under the zero-shot setting.
We then analyze the benefit of the proposed multilingual multimodal pre-training.
%and analyze the translate-train performance.

The upper section shows the zero-shot results.
Unlike cross-lingual transfer in NLP tasks, employing multilingual Transformers in vision-language tasks apparently does not generalize well across languages. 
For example, there is a significant drop in R@1 (19.9$\rightarrow$11.1 (-44\%) with mBERT, 21.0$\rightarrow$16.3 (-24\%) with XLM-R) when directly applying English-finetuned model to German$\rightarrow$video search.
For comparison, there is only a -10\% degradation for XLM-R on $en\rightarrow de$  cross-lingual transfer in XNLI~\cite{conneau2018xnli}.
Multimodal (English-video) pre-training (MP) on HowTo100M only improves average R@1 (+0.1 or mBERT and +1.1 for XLM-R) compared to model-from-scratch.
%
%Results shows that the proposed multilingual multimodal pre-training (MMP) is the key facilitator for zero-shot cross-lingual transfer.
%MMP improves German$\rightarro$Video search (11.1$\rightarrow$15.0, +35\% for mBERT, and 16.3$\rightarrow$19.4, +20\% for XLM-R) and . 
In contrast, our proposed multilingual multimodal pre-training (MMP) is shown to be the key facilitator for zero-shot cross-lingual transfer.
MMP improves German$\rightarrow$Video search (11.1$\rightarrow$15.0, +35\% for mBERT, and 16.3$\rightarrow$19.4, +20\% for XLM-R) and achieves $2.6 \sim 2.8$ improvement in average R@1. 
We attribute the effectiveness of MMP to learning improved alignments between multilingual textual and visual context in the shared embedding space, as relatively balanced improvements between English$\rightarrow$video and non-English$\rightarrow$video is observed with fine-tuning.

% Remark: 1) consider adding degrade percentage from English to other non-English languages and 2) quantify and compare the difference in NLP and V-L models here  

Fig.~\ref{fig:trend} demonstrates the trend of R@1 while incrementally incorporating additional languages for MMP.
For XLM-R, the improvement in R@1 asymptotically converges when pre-training with more multilingual text-video pairs.
On the other hand, for zero-shot German$\rightarrow$video search, pre-training with more languages keeps improving the search performance, even though the additional language (\eg, French) is different from the target language (\ie, German). 
%For instance, the additional French-video pairs for pre-training also benefit zero-shot German-video search.

\begin{figure*}[t!]
    \centering
    \includegraphics[width=1.0\linewidth]{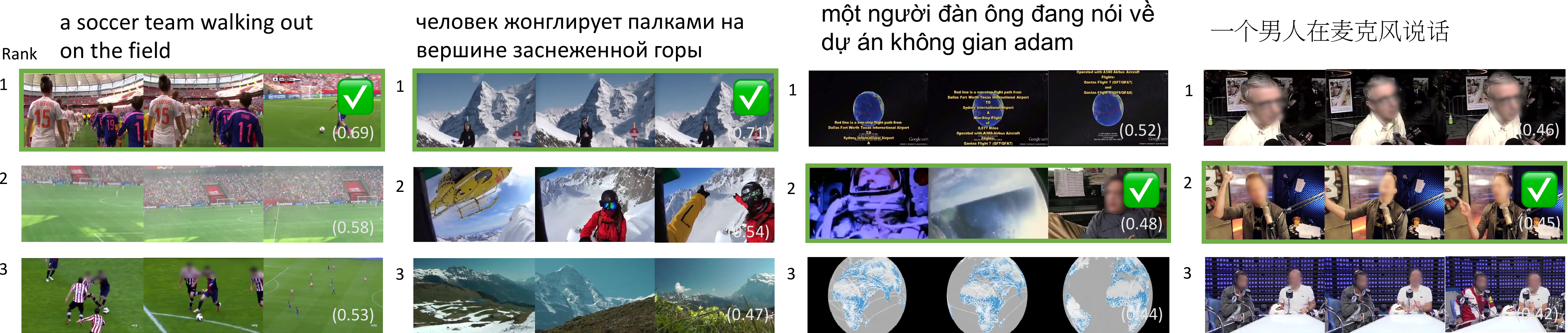}
    \caption{Qualitative multilingual (\textit{en}, \textit{ru}, \textit{vi}, \textit{zh}) text$\rightarrow$video search results on VTT.}\label{fig:qual}
\end{figure*}

The lower section of Table~\ref{tab:cross_lingual} shows the results of models fine-tuned with (synthesized) pseudo-multilingual annotations.
It can be regarded as the \textit{translate-train} scenario, 
%which generally improves upon single language training under the assumption that powerful machine translation model is available
%which serves as the approximate upper bound,
which serves as a strong performance target for evaluating zero-shot cross-lingual transfer, 
as discussed in~\citep{lample-conneau,xtreme}.
Both mBERT and XLM-R yield better performance across non-English languages with the in-domain translated pseudo-multilingual annotations.
However, for English$\rightarrow$video search, a $0.7$ degradation is observed compared to the zero-shot setting. 
It is likely due to the noise in the translated captions.
%Instead of training with uniformly distributed $(s,v,y)$ pairs, we hypothesize that advanced sampling strategies such as exponential sampling in mBERT are likely to alleviate bias towards certain languages and we leave it as the future work.
Notably, there is still a performance gap between zero-shot and translate-train settings for models with mBERT. In contrast, the gap is much smaller for models with XLM-R.
In the following sections, we refer \texttt{Ours-MMP} as our best model with XLM-R as the text backbone and compare it with other state-of-the-art methods.

\noindent \textbf{Qualitative Results}
Fig.~\ref{fig:qual} shows the multilingual text$\rightarrow$video search results with Ours-MMP (VTT:\textit{en}-only) on VTT under the zero-shot setup. 
Note that only one shared English-finetuned model is used for text$\rightarrow$video search in all languages.
As demonstrated, the proposed model successfully retrieves the correct videos with English (\textit{en}) and Russian (\textit{ru}) queries.
The other top-ranked videos also share similar visual appearance to the correct one.
For zero-shot transferring of the English-finetuned model to distant languages such as Vietnamese (\textit{vi}) and Chinese (\textit{zh}), we observe that there is still limitation for our zero-shot models to understand abstract concepts (\eg, ``space project'') and associate small objects (\eg, ``microphone'') with the text queries in distant languages.

\subsection{Comparison to Supervised State of the Art\label{sec:exp:sota}}

\begin{table}[t!]
\small
\setlength{\tabcolsep}{2.0pt}
  \centering
  \begin{tabular}{l ccc}\hline 
  Model  & R@1$\uparrow$ & R@5$\uparrow$  & R@10$\uparrow$ \\
\hline \hline
        JSFusion~\citep{Yu_2018} & $10.2$ & $31.2$ & $43.2$  \\
        JPoSE~\citep{wray2019fine} & $14.3$ & $38.1$ &  $53.0$  \\
        VidTrans\textsuperscript{$\dagger$}~\citep{korbar2020video}   & $14.7$ & $-$ & $52.8$   \\
        HT100M\textsuperscript{$\dagger$}~\citep{ht100m}   & $14.9$ & $40.2$ & $52.8$  \\
        Noise\textsuperscript{$\dagger$}~\citep{amrani2020noise} & $17.4$ & $41.6$ & $53.6$   \\
        CE\footnotemark[2]~\citep{liu2019use}  & $20.9$ & $48.8$ & $62.4$ \\
\hline
        Ours(VTT:$en$-only) & 21.0 & 50.6 & 63.6 \\
        Ours-MMP (VTT:$en$-only) & \textbf{23.8} & \textbf{52.6} & \textbf{65.0} \\
\hline
  \end{tabular}
  \caption{English$\rightarrow$video search performance on VTT. $\dagger$: Models with pre-training on HowTo100M.
  }\label{tab:vtt}
\end{table}

%\subsection{Comparison with SoTA English-Video Models on VTT\label{sec:exp:sota:vtt}}
\noindent \textbf{English$\rightarrow$Video Search on VTT. \label{sec:exp:sota:vtt}}
% English-VTT
Table~\ref{tab:vtt} shows the comparison of English$\rightarrow$video models on VTT.
For a fair comparison to other baselines, our model fine-tunes only with the original English annotations on VTT.
The results show that our model outperforms other baselines by a large margin. 
Specifically, our model achieves 8.9 R@1 improvement over the original HowTo100M model~\cite{ht100m} and other recent baselines with pre-training on HowTo100M. 
Using a smaller set of visual features and training on a smaller (6,513 vs 9,000) training set\footnote{CE uses 9,000 videos (VTT training and part of exclusive testing set) for training, while other baselines and our model in Table~\ref{tab:vtt} are trained on the official VTT training set which contains 6,513 videos.}, our model also outperforms CE~\cite{liu2019use} with or without pre-training.

\begin{table}[t!]
\small
\centering
\resizebox{0.48\textwidth}{!}{%
\setlength{\tabcolsep}{2.0pt}
\begin{tabular}{lccccccccc}
\hline
 & \multicolumn{3}{c}{English to Video} & \multicolumn{3}{c}{Chinese to Video} \\
Model & R@1$\uparrow$ & R@5$\uparrow$ & R10$\uparrow$ & R@1$\uparrow$ & R@5$\uparrow$ & R@10$\uparrow$  \\ \hline\hline
VSE~\cite{vse} & 28.0 & 64.3 & 76.9 & - & - & - \\
VSE++~\cite{faghri2018vse++} & 33.7 & 70.1 & 81.0  & - & - & - \\ 
Dual~\cite{dual} & 31.1 & 67.4 & 78.9 &  - & - & - \\
HGR~\cite{hgr} & 35.1 & 73.5 & 83.5 & - & - & - \\
\hline
Ours (VATEX:$en$-only) & 43.5 & 79.8 & 88.1 & 23.9 & 55.1 & 67.8 \\
Ours-MMP (VATEX:$en$-only) & \textbf{44.4} & 80.5 & 88.7 & 29.7 & 63.2 & 75.5 \\
Ours-MMP (VATEX:$en,zh$) & 44.3 & \textbf{80.7} & \textbf{88.9} & \textbf{40.5} & \textbf{76.4} & \textbf{85.9} \\
\hline
\end{tabular}}
\caption{Multilingual text$\rightarrow$video search on VATEX.} \label{tab:vatex}
\vspace{-1.0em}
\end{table}
% remark: consider translate vatex to 10 langs? (translate-test 

\noindent \textbf{Multilingual Text$\rightarrow$Video Search on VATEX.\label{sec:exp:sota:vatex}}
% Cross-lingual VATEX
Table~\ref{tab:vatex} summarizes English$\rightarrow$video and Chinese$\rightarrow$video search performance on the VATEX dataset.
%As observed, our model generalizes well across distant languages (English and Chinese).
Under the zero-shot setting where we train with only English-video pairs, our model already outperforms other baselines. 
However, a clear performance gap between English$\rightarrow$video and Chinese$\rightarrow$video search is observed, indicating that cross-lingual transfer to a distant language remains challenging even with XLM-R.
With the proposed MMP, the gap is significantly closed by 5.8/8.1/7.7 in R@1/5/10.
When in-domain human-annotated Chinese captions are available, 
the performance of our model can further be improved for both languages and our model yields new state-of-the-art performance.

\begin{table*}[t!]
\small
\centering
\setlength\tabcolsep{3.0pt}
\begin{tabular}{lrcccccccccccc}
\hline
 & M30K & \multicolumn{3}{c}{English to Image} & \multicolumn{3}{c}{German to Image} & \multicolumn{3}{c}{Czech to Image} \\
Model & \# lang. & R@1$\uparrow$ & R@5$\uparrow$ & R10$\uparrow$ & R@1$\uparrow$ & R@5$\uparrow$ & R@10$\uparrow$ & R@1$\uparrow$ & R@5$\uparrow$ & R@10$\uparrow$  \\ \hline\hline
OE~\cite{vendrov2015order} & 2 & 21.0 & 48.5 & 60.4  & 25.8 & 56.5 & 67.8 & - & - & -\\ 
VSE++~\cite{faghri2018vse++} & 2 & 31.3 & 62.2 & 70.9  & 39.6 & 69.1 & 79.8 & - & - & -\\ 
Pivot~\cite{gella2017image} & 2 & 22.5 & 49.3 & 61.7 &  26.2 & 56.4 & 68.4 & - & - & - \\
FB-NMT~\cite{forward} & 2 & 47.3 & 75.4 & 83.5 & 37.0 & 64.0 & 73.1 & - & - & - \\
MULE~\cite{mule} & 4 & 42.2 & 72.2 & 81.8 & 35.1 & 64.6 & 75.3 & 37.5 &64.6 & 74.8 \\
SMALR~\cite{smalr} & 10 & 41.8 & 72.4 & 82.1 & 36.9 & 65.4 & 75.4 & 36.7 & 68.0 & 78.2 \\
MHA-D~\cite{mhad} & 2 & 50.1 & 78.1 & 85.7 & 40.3 & 70.1 & 79.0 & - & - & - \\ 
\hline
Ours (M30K:$en$-only) & 1 & 48.4 & 78.3 & 85.9 & 31.4 & 61.1 & 72.6 & 33.2 & 65.2 & 76.1 \\
Ours-MMP (M30K:$en$-only) & 1 & 50.0 & 79.2 & 86.8 & 33.8 & 63.3 & 74.7 & 37.9 & 68.8 & 78.2 \\
Ours-MMP (M30K:$en,de,cs,fr$) & 4 & \textbf{51.6} & \textbf{80.1} & \textbf{87.3} &  \textbf{45.1} & \textbf{75.6} & \textbf{85.0} & \textbf{46.6} & \textbf{75.9} & \textbf{83.4} \\
\hline
\end{tabular}
\caption{Multilingual text$\rightarrow$image search on Multi30K. MMP: Multilingual multimodal pre-training.} \label{tab:m30k}
\vspace{-1.0em}
\end{table*}

\noindent \textbf{Cross-Modality Transfer to Multi30K: From Video-Text to Image-Text\label{sec:exp:img_text}.}
To extend our study on zero-shot cross-lingual transfer for image-text tasks, we investigate the feasibility of transferring our video-text model across modalities.
We replace the 3D-CNN in the original video-text model with a 2D-CNN to encode the image. 
In practice, following MHA-D~\citep{mhad},
we utilize the Faster-RCNN~\cite{ren2015faster} pre-trained in Visual Genome~\cite{krishnavisualgenome} to extract regional visual features.
Essentially, an image is encoded as $e_v=\mathbb{R}^{M \times H}$ where $M=36$ is the maximum number of visual objects in an image.
For models with MMP, we initialize their weights with the model pre-trained on Multi-HowTo100M.
To tackle the feature mismatch between 2D-CNN and 3D-CNN, we leverage a linear layer with a doubled learning rate to map 2D-CNN features to the same dimension as 3D-CNN features.

Table~\ref{tab:m30k} shows the results on Multi30K. 
For zero-shot cross-lingual transfer, when trained from scratch (M30K:$en$-only), our model achieves comparable performance to MHA-D but lags in German$\rightarrow$image search since it only uses English annotations.
In Ours-MMP, pre-training improves all recall metrics even with modality gap. The average R@1 improvement is 3.2.
A larger gain for (relatively) low-resource language such as Czech is observed.
Without using any Czech annotations, our zero-shot model with MMP achieves comparable Czech$\rightarrow$image search performance to SMALR~\cite{smalr}, which uses 10 languages including Czech.
However, when transferring across modalities and using only English annotations, there are performance gaps between English$\rightarrow$Image and German/Czech$\rightarrow$Image search, implying that transferring models across modalities is feasible but remains challenging. 
We consider zero-shot cross-modal cross-lingual transfer 
%of varieties of vision-language models (\eg~VQA~\cite{balanced_vqa_v2})
as our future work.

For a fair comparison with other baselines, when trained with annotations in all 4 languages provided by Multi30K, 
our model greatly outperforms all baselines by large margins in multilingual text$\rightarrow$image search. 
%Notably, English$\rightarrow$image search performance can further be improved by training with multilingual annotations.

\section{Conclusion}

%\gn{This conclusion doesn't add very much to the paper, and mostly just repeats what people who have read the paper will already know.. You could almost say ``The results in this paper have convincingly demonstrated that multilingual multimodal pre-training is an essential ingredient of cross-lingual transfer of vision-language models. We believe our proposed MMP methodology, and corresponding resources we will release, will be a first step towards spurring more research in this direction. However, there are many remaining challenges, such as XXX.''}

%We have posited a rewarding challenge that transfers vision-language models across languages. 
We have presented a multilingual multimodal pre-training (MMP) strategy, the Multi-HowTo100M dataset, and a Transformer-based text-video model for learning contextual multilingual multimodal representations.
The results in this paper have convincingly demonstrated that MMP is an essential ingredient for zero-shot cross-lingual transfer of vision-language models.
Meanwhile, there are many remaining challenges, such as resolving the performance gap between zero-shot and training with in-domain non-English annotations; 
as well as techniques to transfer varieties of vision-language models (\eg, VQA~\cite{balanced_vqa_v2}, TVQA~\cite{lei-etal-2020-tvqa}) or visually-enhanced NLP models such as unsupervised multimodal machine translation~\cite{ummt}. 
We believe the proposed methodology, and the corresponding resources we release, will be an important first step towards spurring more research in this direction.

%We have presented a multilingual multimodal pre-training (MMP) strategy and a Transformer-based text-video model for learning contextual multilingual multimodal representations.
%Empirical results show that MMP facilitates zero-shot cross-lingual transfer of video-text models.
%In addition, MMP improves English-Video search and our method achieves SoTA in multilingual video search on VTT and VATEX.
%Furthermore, we showcase the feasibility of transferring pre-trained video-text models to image-text models on Multi30K, where our method also achieves state of the art in multilingual image search.

%\newpage

\section*{Acknowledgments}
This work is supported by the DARPA grants funded under the AIDA program (FA8750-18-2-0018) and the GAILA program (award HR00111990063) (P.Y.).
This work is also supported by EPSRC Centre for Doctoral Training in Autonomous Intelligent Machines \& Systems [EP/L015897/1] (M.P.).
The authors appreciate Prahal Arora, Shengxin Zha, Polina Kuznetsova, Xu Hu, and Geoffrey Zweig for their suggestions of this work. 
The authors would also like to thank the anonymous reviewers for their feedback.

\newpage
%\balance
%\bibliography{anthology,acl2020,refs}
\bibliography{acl2020,refs}
\bibliographystyle{acl_natbib}

\newpage
\appendix

\section{Appendix Overview}

The Appendix is organized as follows: 
First we provide details about the Multilingual HowTo100M (Multi-HowTo100M) dataset for multilingual multimodal pre-training (MMP) in \secref{app:how2}. 
Then we provide additional implementation details and experiment setup in \secref{app:exp:setup}. 
Additional ablation studies regarding choices of Transformer architecture are discussed in \secref{app:hyper}. 
Then we present additional cross-dataset transfer experiments in \secref{app:exp:results}. 
%Additional qualitative results on VTT can be found in \secref{app:exp:qual}.

\section{The Multilingual HowTo100M Dataset\label{app:how2}}
In this section we provide the detailed statistics of the Multilingual HowTo100M (Multi-HowTo100M) dataset. 
We also provide a comparison to ~\citet{sigurdsson2020visual} that also uses HowTo100M for unsupervised word translation.

The Multi-HowTo100M dataset is built upon the original English HowTo100M dataset~\cite{ht100m} that contains 1.2 million instructional videos (138 million clips) on YouTube. 
We reuse the \emph{raw} English subtitles in HowTo100M, where the subtitles in HowTo100M are either automatic speech recognition (ASR) transcriptions or user generated subtitles.

For Multi-HowTo100M, we use the same video collection as English HowTo100M.
At the time of data collection (May 2020), there were 1.09 million videos accessible.
We collect the subtitles provided by YouTube, which either consist of user-generated subtitles or those generated by Google ASR and Translate in the absence of user-generated ones.
Essentially, we collect video subtitles in 9 languages: English (\textit{en}), German (\textit{de}), French (\textit{fr}), Russian (\textit{ru}), Spanish (\textit{es}), Czech (\textit{cz}), Swahili (\textit{sw}), Chinese (\textit{zh}), Vietnamese (\textit{vi}). 
Table~\ref{app:stat} summarizes the dataset statistics for each language. In most cases there are more than 1 billion tokens a language.

\begin{figure}[t!]
     \centering
     \centering
     \includegraphics[width=0.49\textwidth]{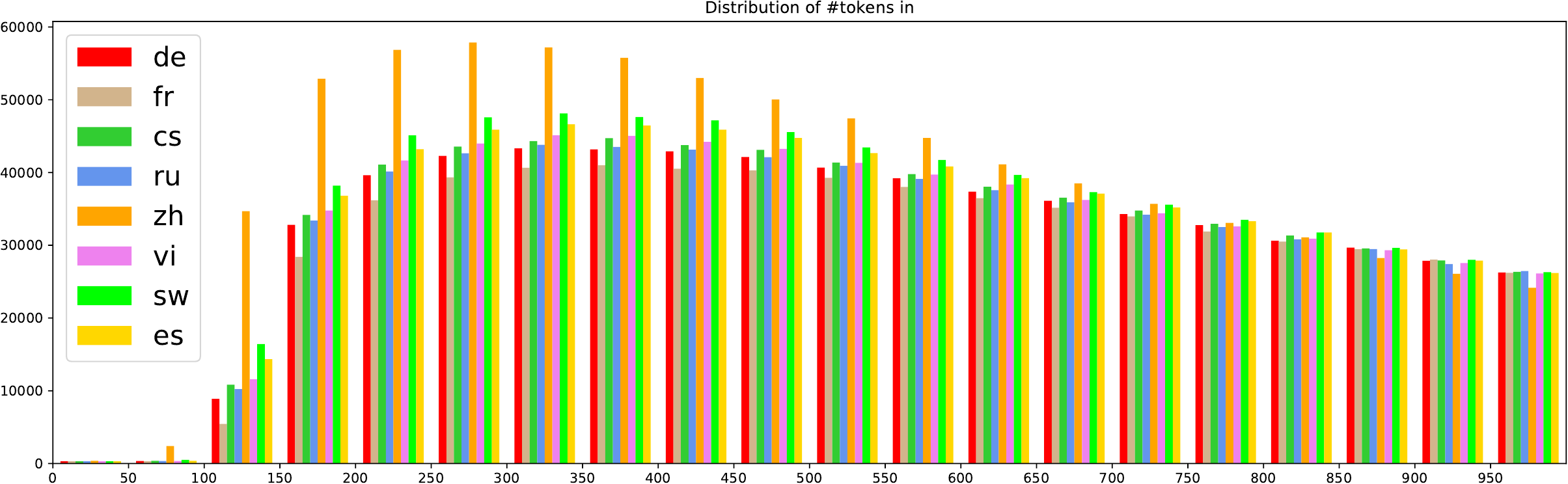}
     \caption{Distribution of \#tokens/video in Multi-HowTo100M
     }
     \label{fig:app:num_tok}
 \end{figure}

\begin{figure}[t!]
     \centering
     \centering
     \includegraphics[width=0.49\textwidth]{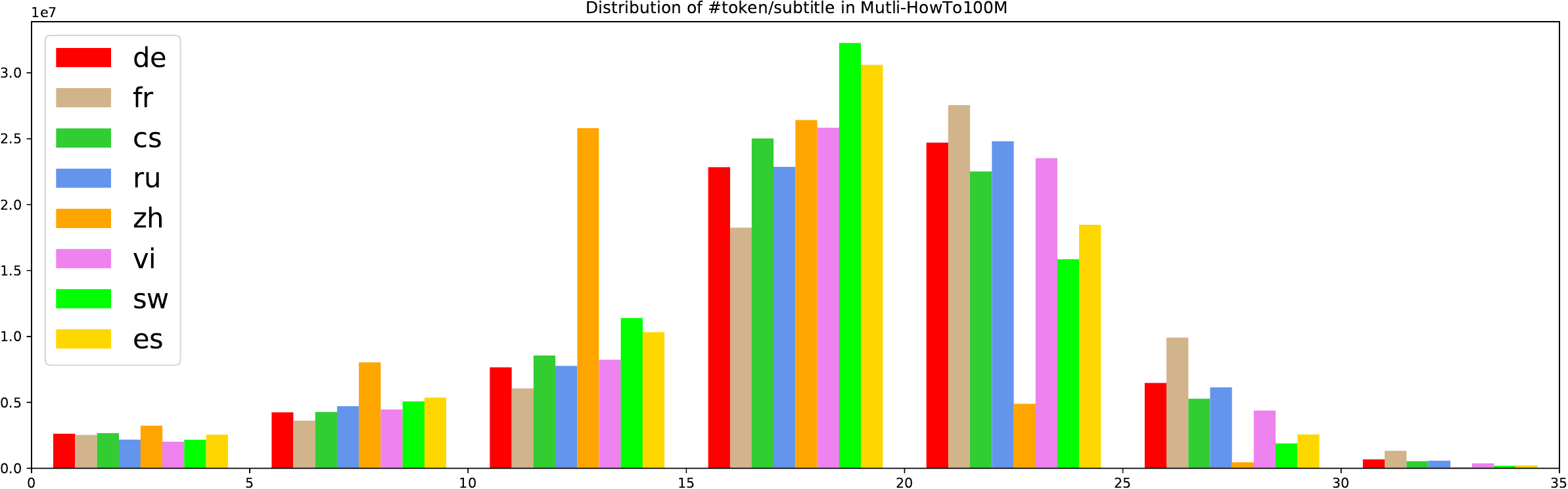}
     \caption{Distribution of \#tokens/subtitle in Multi-HowTo100M
     }
     \label{fig:app:num_sub}
\end{figure}

Fig.~\ref{fig:app:num_tok} further shows the number of tokens per video. 
There are typically lengthy narrations that contains several hundreds of tokens available in each instructional video. 
Fig.~\ref{fig:app:num_sub} shows the distribution of number of tokens in a subtitle. 
For each subtitle segment, which ranges from 0$\sim$20 seconds, there are typically 15$\sim$25 words. 
The most of the cases, subtitles are well aligned in time for non-English languages. 
Fig.~\ref{fig:dataset_vis} visualizes a few examples in Multi-HowTo100M.

\begin{table}[t!]
\setlength{\tabcolsep}{1.5pt}
  \centering
  \begin{tabular}{l ccc}\hline 
    Language & videos & \#subtitle &\#tokens \\
\hline \hline
    English & 1238911 & 138429877 & 1.18B\\
    German & 1092947 & 69317890& 1.26B\\
    French & 1093070 & 69399097 & 1.33B\\
    Czech & 1092717 & 68911940 & 1.22B\\
    Russian & 1092802 & 69117193& 1.25B\\
    Chinese & 1092915 & 68939488& 0.94B\\
    Swahili & 1092302 & 68898800 & 1.22B\\
    Vietnamese & 1092603 & 68887868& 1.13B\\
    Spanish & 1092649 & 70143503 &  1.16B\\
\hline
  \end{tabular}
  \caption{Multi-HowTo100M statistics\label{app:stat}}
\end{table}

A similar HowTo100M variant has been recently reported in MUVE~\citep{sigurdsson2020visual} that is created for unsupervised word translation. 
Our Multi-HowTo100M differs from MUVE in the following perspectives: 
First, we collects 9 language for \emph{all} videos in HowTo100M while MUVE only has 4 languages available (English, French, Japanese, and Korean) on HowTo100M. 
Also, MUVE divided HowTo100M into 4 non-overlapped sections for each language, there are no parallel pairs for each subtitle.
While in Multi-HowTo100M, there are 7-9 languages for each subtitle. 
Essentially, There are more than 1 billion tokens in most languages in Multi-HowTo100M.
To our best knowledge, our Multi-HowTo100M dataset is currently the largest multilingual text-video collection.

Beyond scale, instructional videos in Multi-HowTo100M are feasible pre-training resources for many downstream vision-language models.
Demonstrators in instructional videos typically perform intentionally and explain the visual object or action explicitly.
According to the inspection by~\citep{ht100m}, for around 51\% of clips, at least one object or action mention in the caption can be visually seen.
Prior work has shown that instructional videos are useful for event recognition~\cite{iyu}, action localization model~\cite{Alayrac16unsupervised}, cross-modal alignments~\cite{malmaud-etal-2015-whats}.
We expect the previous success in the intersection of natural language processing (NLP) and computer vision (CV) could be further translated into more languages to have a broaden impact.

The are great potentials of using our Multi-HowTo100M dataset in related research field such as multilingual multimodal representation learning~\cite{mhad,mule,smalr}, multilingual multimodal translation~\cite{ummt,Globetrotter}, multilingual image/video captioning~\cite{cocojap} ... etc.\ We expect the release of Multi-HowTo100M will be a first step towards spurring more research in these directions.

\section{Implementation and Experiment Details\label{app:exp:setup}}

\paragraph{Pre-Processing.}
For pre-possessing, we truncate the maximum length $N$ of text to 192 for pre-training on Multi-HowTo100M. 
The maximum length is set to 96 for fine-tuning VTT~\cite{vtt}, VATEX~\cite{vatex} and Multi30K~\cite{Multi30K}.
The maximum video length $M$ is set to 128 for pre-training on Multi-HowTo100M and 36 for all fine-tuning tasks.

\paragraph{Model Architecture.}
For the multilingual Transformers, either multilingual BERT~\cite{devlin-etal-2019-bert} or XLM-R-large~\cite{artetxe-etal-2020-cross}, we use the pre-trained version provided by HuggingFace.~\footnote{https://github.com/huggingface/transformers} and use their corresponding tokenizers for tokenization. Detailed design choices regarding output layer and frozen layer is discussed in~\secref{app:hyper}.

For the video backbone, we use a 34-layer, R(2+1)-D~\citep{tran2018closer} network pre-trained on IG65M~\cite{ghadiyaram2019large} and a S3D~\cite{miech19endtoend} network pre-trained on HowTo100M~\cite{ht100m}.
We apply a spatial-temporal average pooling over the last convolutional layer, resulting in a 512-dimensional vector for each 3D CNN network.
We extract visual features at a rate of 1 feature per second.
Since the 3D CNNs employs different size of input windows (\eg, 8 frames for R(2+1)D and 16 for S3D), we re-sample videos to 30 fps and employs a window of size 8 or 30 that takes consecutive frames starting from the beginning of every second for encoding.
We simply concatenate the two 3D-CNN outputs and use the 1024-dimension vector as the visual input stream to our model.
Notably, instead of using 9 different types of visual features as in CE \citep{liu2019use}, we use only the above 2 features and achieve superior performance.

For the Transformer pooling head (TP) modules, we use a 2-layer Transformer with 4-head attention for each TP.
The embedding dimension $D$ is set to 1024.
We do not use the positional embeddings in both text and video TP as we do not find them beneficial in our experiments.
The softmax temperature in all NCE contrastive objectives is set to $0.1$ as used in SimCLR~\cite{Chen2020ASF}.

Note that unlike ViLBERT~\cite{vilbert} or OAN~\cite{oan}, our models does not employ cross-modality attention and keep the multi-head self-attention within the same modality.
The main reason is to reduce the inference time complexity.
For cross-modality attention, the complexity is $O(TV)$ to encode $T$ text queries for $V$ videos in a dataset before retrieval (since video and query representations depend on each other). 
It is clearly not scalable when the dataset contains millions of videos. 
To this end, our model keep self-attention within the same modality which results in a $O(T+V)$ complexity compared $O(TV)$ in prior work with cross-modality attention.
In our preliminary experiments, we also incorporate cross-modality attention and achieved 0.3$\sim$1.8 R@1 improvement. Considering the trade-off between performance and scalability, we choose the latter.

\paragraph{Training and Inference Details and Profiling.}
For the softmax temperature in NCE, we set to 0.1 as used in SimCLR~\cite{Chen2020ASF}. We use the Adam~\citep{kingma2014adam} optimizer with a initial learning rate $2\cdot 10^{-4}$ and clip gradients greater than 0.2 during the training phase. 
Dropout rate is 0.3. 
Since the video length and token length is longer in the pre-training phase, we use a 64 batch size for pre-training. For fine-tuning, we use a batch size of 128.

Pre-training on the 1.2 million HowTo100M videos takes around 10 GPU hours (NVIDA V100) for 16 epochs.
We speed up the pre-training process by distributing the workload over 8 GPUs on a single node of our server. 
We use 1 GPU for the fine-tuning or training from scratch experiments.
For the MSR-VTT split, it takes 12 GPU hours to train our model on 180K video-text pairs for 20 epochs. 
For VATEX, it takes 32 GPU hours to train on 260K video-text pairs for 30 epochs. 
For inference, the encoding speed is around 250-300 videos/sec and 200-250 text queries/sec. 
The overall text$\rightarrow$video search  speed on 1,000 video-text pairs (1,000 text queries over 1,000 videos) is around 6 seconds including video/text encoding and ranking  their similarity scores.

\paragraph{Experiment Details.}
Our experiment consider three types of pre-training: (1) Multilingual multimodal pre-training (MMP), (2) Multimodal pre-training (MP), and (3) no pre-training (from scratch).
For (1) and (2), we pre-train 16 epochs and use the model weight at 16-th epoch for fine-tuning experiments.

For multimodal pre-training, we pre-train on the original English HowTo100M dataset. We iterate over all videos in HowTo100M. For each video, we randomly sample the start and end time to construct a video clip. For each clip, we locate the nearest consecutive ASR transcriptions in time and use it as to construct the (video, text) pair for training.

For multilingual multimodal pre-training (MMP), we use Multi-HowTo100M for pre-training.
For each video, we follow the same strategy as MP.
For a clip, we sample one language type each time from 9 languages and use the consecutive ASR transcriptions that are closest in time to compose (video, text) pairs for training.

After pre-training, we fine-tune our model on VTT and VATEX to evaluate on text$\rightarrow$video search tasks. 
In the zero-shot cross-lingual transfer experiments, we use only English-video data. 
We then directly test the model with non-English queries to report the zero-shot performance.
When annotations in additional languages are available
(by humans in VATEX and Multi30K; by MT models (\ie~\textit{translate-train}) in VTT),
we train our model with all available multilingual annotations (\ie~fully supervised) to compare fairly with other baselines in multilingual text$\rightarrow$video search.

Since pre-trained model has a faster convergence rate, we fine-tune for 10 epochs and use the model with best validation performance (summation of R@1, R@5, R@10) for testing.
For models without pre-training (\ie, from-scratch), we train for 20 epochs under the same training protocol.

\begin{table}[t!]
\small
\centering
    \setlength{\tabcolsep}{1.5pt}
      \begin{tabular}{ccccc}\hline
      Output layer  & Freeze lower &\textit{en} & \textit{de}\\
      \hline \hline
        3	& 0  & 20.9 & 3.2 \\
        6	& 0  & 20.5 & 3.1 \\
        9	& 0  & 21.0 & 4.8 \\
        12	& 0  & 21.0 & 13.3 \\ 
        15	& 0  & 20.5 & 12.3 \\ 
        18	& 0  & 20.8 & 12.6 \\ 
        \hline 
        12 & 6 & 21.0 & 15.5 \\
        12 & 9 & \textbf{21.0} & \textbf{16.3} \\
        12 & 12 & 18.9 & 14.1\\
      \hline
    \end{tabular}
    \caption{Text$\rightarrow$video R@1 of XLM-R output layers and layers to freeze on VTT}\label{tab:supp:xlm_layer}
\end{table}

\begin{table}[t!]
\small
\centering
    \setlength{\tabcolsep}{1.5pt}
      \begin{tabular}{ccccc}\hline
      Output layer  & Freeze lower &\textit{en} & \textit{de} \\
      \hline \hline
        3	& 0  & 19.2 & 2.5 \\
        6	& 0  & 19.5 & 2.0 \\
        9	& 0  & 19.3 & 5.8 \\
        12	& 0  & 19.6 & 8.8 \\ 
        \hline 
        12 & 6 & 19.3 & 10.5 \\
        12 & 9 & \textbf{19.9} & \textbf{11.1} \\
        12 & 12 & 18.9 & 9.8\\
      \hline
    \end{tabular}
    \caption{Text$\rightarrow$video R@1 of mBERT output layers and layers to freeze on VTT}\label{tab:supp:mbert_layer}
\end{table}

\section{Additional Ablation Studies\label{app:hyper}}
As has been investigated in XTREME~\cite{xtreme}, choosing different output layers will affect the zero-shot transferability of multilingual Transformers in various NLP tasks. 
For text$\rightarrow$video search tasks, we conduct a series of experiments to identify the desirable choices of hyper-parameters in the proposed multilingual multimodal Transformer that lead to best performance in English-to-video and (zero-shot) non-English-to-video search performance. 
Beyond our ablation studies in Sec. 5, in this part we highlight our trials in the choice of the output layer and the layers to be frozen in our multilingual Transformer backbone (\ie, mBERT and XLM-R).
There are 24 layers in XLM-R (large) and 12 layers in mBERT.
We perform grid-search on VTT to identify the best choice of these two hyper-parameters. 

\paragraph{Choice of Output Layers}
Table~\ref{tab:supp:xlm_layer} and Table~\ref{tab:supp:mbert_layer} compare different choices of output layer and layers to freeze in multilingual Transformers.
Our results suggest that the best output layer for mBERT and XLM-R is the 12-th layer. 
Surprisingly, while output layer does not affect English$\rightarrow$video search significantly, it greatly affects the zero-shot cross-lingual transfer performance of video-text models.
For both XLM-R and mBERT, the performance degrade significantly if fine-tuning all layers.

\paragraph{Choice of Layers to Freeze}
Similar to output layers, the choice of frozen layers greatly affects cross-lingual transferability. For both mBERT and XLM-R, it is desirable to freeze part of the lower layers and make the top-3 layers trainable for video-text models.
We observe that when freezing all layers (\ie, using the pre-extracted contextual multilingual embeddings) does not lead to satisfactory results. 
For mBERT, R@1 drops from $19.9$ to $18.9$ in English$\rightarrow$video search and $11.1$ to $9.8$ in German$\rightarrow$video search.
For XLM-R, R@1 drops from $21.0$ to $18.9$ in English$\rightarrow$video search and $16.3$ to $14.1$ in German$\rightarrow$video search.
These results imply that text-only contextual multilingual embeddings along are likely to be infeasible to be applied to vision-language tasks without proper fine-tuning.

An important observation is that the best English$\rightarrow$video search performance corresponds to the best German$\rightarrow$video performance.
This trend implies that for model selection, the configuration for the best English$\rightarrow$video model usually translates to the best configuration for (zero-shot) cross-lingual model. 
This shared trend justifies the English$\rightarrow$video ablation studies in the original paper. 
Note that we utilize the best English$\rightarrow$video for all (zero-shot) cross-lingual experiment in our experiment section.

For multilingual text$\rightarrow$video search, the best configuration we found in our experiments is to output the 12-th layer and freeze the layers below 9 for both mBERT and XLM-R.

%\begin{figure*}[t!]
%     \centering
%     \includegraphics[width=\textwidth]{figs/multi_qual2_.pdf}
%     \label{fig:vtt_qual2}
%     \caption{Qualitative examples of the top-3 multilingual text$\rightarrow$video search results and cosine similarity scores on VTT. Only one correct video (colored in green) for each multilingual text query on the top.
%     }
%     \label{fig:app:qual}
%\end{figure*}

\begin{table}[t!]
\small
\centering
      \begin{tabular}{ccc}\hline
        text$\rightarrow$video & English & Non-English \\
      \hline \hline
      In-domain & \checkmark & \checkmark\\
      Out-of-domain & \checkmark & \\
      \hline
    \end{tabular}
    \caption{Coverage of our experiments}\label{tab:supp:scope}
\end{table}

\section{Additional Experimental Results\label{app:exp:results}}
The coverage of our text$\rightarrow$video search experiments is summarized in Table~\ref{tab:supp:scope}. Our experiments cover the following scenarios: \\
\noindent\textbf{In-domain, English}: Table 5 (VTT) and Table 6 (VATEX) in the original paper.\\
\noindent\textbf{In-domain, non-English}: Table 4 (VTT, 9 languages) and Table 6 (VATEX, Chinese).\\
\noindent\textbf{Out-of-domain, English}: Additional (zero-shot) generalization results across datasets are in \secref{sec:exp:cross_dataset}.\\
\noindent\textbf{Out-of-domain, non-English}: We consider this as our future work.

\begin{table}[t!]
\small
\centering
\setlength{\tabcolsep}{1.5pt}
\begin{tabular}{lccc}\hline
Model & R@1 & R@5& R@10  \\
\hline\hline
VSE~\cite{vse} & 10.1 & 29.4 & 41.5   \\
VSE++~\cite{faghri2018vse++} & 14.4 & 35.7 & 46.9  \\
Dual~\cite{dual}   & 13.7 & 36.1 & 48.2  \\
HGR~\cite{hgr} & 16.4 & 38.3 & 49.8  \\ \hline
\textbf{Ours}-Full & \textbf{24.0} & \textbf{50.5} & \textbf{62.1} \\ \hline
\end{tabular}
\caption{Zero-shot generalization on YouTube2Text with VTT-finetuned model.}
\label{tab:y2t}
\end{table}

\subsection{Generalizability across English-Video Datasets\label{sec:exp:cross_dataset}}

In this section. we provide additional experiment results regarding zero-shot generalization of the VTT-finetuned model on 
out-of-domain dataset.
Specifically, we test on YouTube2Text~\cite{youtube2text}.
The aim of this experiment is to test the cross-dataset generalizabilty of our model without using domain-specific training data.

Table~\ref{tab:y2t} shows the comparison of English$\rightarrow$video search results on the YouTube2Text testing set. 
Models in this table are only fine-tuned on VTT and use \textit{no} YouTube2Text training data.
As can be observed, our model with MMP generalizes well on YouTube2Text, outperforming HGR~\cite{hgr} by 7.6 and DualEncoder~\cite{dual} by 10.3 in R@1.

%\section{Additional Qualitative Results\label{app:exp:qual}}
%We provide addition qualitative multilingual text$\rightarrow$video search results on VTT in Fig.~\ref{fig:app:qual}.
%With a query in 6 possible languages, there is one and only one correct video to be retried out of the 1000 testing videos in VTT testing set. 
%As can be observed, given a multilingual text query on top, in most cases, our model successfully retrieves the correct videos marked in green. Also, the top-ranked videos look semantically similar to the correct one.

\end{document}